\documentclass[conference]{IEEEtran}

\ifCLASSINFOpdf
\else
   \usepackage[dvips]{graphicx}
   \graphicspath{{../eps/}}
   \DeclareGraphicsExtensions{.eps}
\fi

\usepackage[utf8]{inputenc}
\usepackage{graphicx,url}
\usepackage{amsmath}
\usepackage[normalem]{ulem}
\usepackage{multirow}
\usepackage{xcolor}

\usepackage{balance}
\usepackage{color}

\usepackage{booktabs}
\usepackage{cite}
\usepackage[acronym]{glossaries}
\usepackage[colorlinks=true,pagebackref=false,urlcolor=black,linkcolor=black,citecolor=black]{hyperref} %
\newacronym{hog}{HOG}{Histogram of Oriented Gradients}
\newacronym{svm}{SVM}{Support Vector Machines}
\newacronym{cnn}{CNN}{Convolutional Neural Network}
\newacronym{iiitdcli}{IIIT-D CLI}{IIIT-Delhi Contact Lens Iris}
\newacronym{ndcld15}{NDCLD15}{Notre Dame Contact Lens Detection 2015}
\newacronym{mobbiofake}{MobBIOfake}{MobBIOfake}
\newacronym{ndccl}{NDCCL}{Notre Dame Cosmetic Contact Lenses}
\newacronym{iou}{IoU}{Intersection over Union}
\ifCLASSOPTIONcompsoc
    \usepackage[caption=false, font=normalsize, labelfont=sf, textfont=sf]{subfig}
\else
	\usepackage[caption=false, font=footnotesize]{subfig}
\fi

\begin{document}

\title{A Benchmark for Iris Location and a Deep Learning Detector Evaluation}

\author{
\IEEEauthorblockN{
Evair Severo\IEEEauthorrefmark{1}, Rayson Laroca\IEEEauthorrefmark{1}, Cides S. Bezerra\IEEEauthorrefmark{1}, Luiz A. Zanlorensi\IEEEauthorrefmark{1},\\
Daniel Weingaertner\IEEEauthorrefmark{1}, Gladston Moreira\IEEEauthorrefmark{2} and David Menotti\IEEEauthorrefmark{1}
}
\IEEEauthorblockA{
\IEEEauthorrefmark{1}%
\textit{Postgraduate Program in Informatics},  
\textit{Federal University of Paran\'a (UFPR)}, 
\textit{Curitiba, Paran\'a, Brazil} \\
\IEEEauthorrefmark{2}%
\textit{Computing Department},  
\textit{Federal University of Ouro Preto (UFOP)}, 
\textit{Ouro Preto, Minas Gerais, Brazil} \\
\\
Email: \{ebsevero, rblsantos, csbezerra, lazjunior, daniel, menotti\}@inf.ufpr.br\ \ \ gladston@iceb.ufop.br
}
}

\maketitle

\begin{abstract}
The iris is considered as the biometric trait with the highest unique probability.
The iris location is an important task for biometrics systems, affecting directly the results obtained in specific applications such as iris recognition, spoofing and contact lenses detection, among others.
This work defines the iris location problem as the delimitation of the smallest squared window that encompasses the iris region.
In order to build a benchmark for iris location we annotate (iris squared bounding boxes) four databases from different biometric applications and make them publicly available to the community.
Besides these 4 annotated databases, we include 2 others from the literature. We perform experiments on these six databases, five obtained with near infra-red sensors and one with visible light sensor.
We compare the classical and outstanding Daugman iris location approach with two window based detectors:
1) a sliding window detector based on features from \gls*{hog} and a linear \gls*{svm} classifier;
2) a deep learning based detector fine-tuned from YOLO object detector.
Experimental results showed that the deep learning based detector outperforms the other ones in terms of accuracy and runtime (GPUs version) and should be chosen whenever possible.
\end{abstract}


\IEEEpeerreviewmaketitle

\vspace{-0.5mm}
\section{Introduction}
\label{sec:intro}

\glsresetall

\begin{figure}[!htb]
	\vspace{-2.5mm}
	\centering
	\subfloat[][\label{fig:periocular}]{%
		\includegraphics[width=0.24\textwidth]{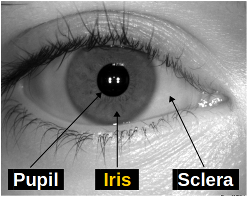}} \hfill
    \subfloat[][\label{fig:iris}]{%
		\includegraphics[width=0.24\textwidth]{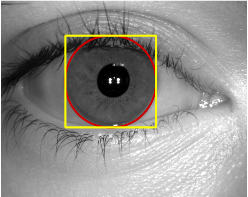}}
	\caption{(a) Periocular region and its main structures. (b) Manual iris location through a bounding box and a circle.}
	\label{fig:eye}   
	\vspace{-3mm} 
\end{figure}

Biometrics systems have significantly improved person identification and authentication, performing an important role in personal, national and global security~\cite{menotti2015deep}.
In biometry, the iris appears as one of the main biological characteristics, since it remains unchanged over time and is unique for each person~\cite{zhu2000biometric}. 
Furthermore, the identification process is non-invasive, in other words, there is no need of physical contact to obtain an iris image and analyze it~\cite{jain2006biometrics}.
Figure~\ref{fig:periocular} illustrates the iris and other structures of a human eye.

Iris location is usually the initial step in recognition, authentication and identification systems~\cite{daugman1993high} and thus can directly affect their performance~\cite{wildes1997iris,daugman2004iris}.
In this sense, how the iris location step influences those systems is an interesting question to be studied.
For achieving such aim, here, we propose to benchmark/evaluate baseline methods that can be applied to iris location.
Initially, we survey some methods in the literature.

The pioneer and maybe the most well known methods for iris location is the one proposed by Daugman~\cite{daugman2004iris}, which defines an integro-differential operator to identify the circular borders present in the images. 
This operator takes into account the circular shape of the iris in order to find the correct position, by maximizing the partial derivative with respect to the radius.

Wildes~\cite{wildes1997iris} proposed another relevant method for iris location by using border detection and the Hough transform. 
First, the iris is isolated by using Gaussian filters of low pass followed by a spatial sub-sampling. 
Subsequently, the Hough transform is applied and those elements that better fit a circle according to a defined condition are selected.

Tisse et al.~\cite{tisse2002iris}, present a modification of Daugman's algorithm. 
This approach applies a Hough transform on a gradient decomposition to find an approximation of the pupil center. 
Then, the integro-differential operator is applied to locate the iris boundaries.
It has the advantage of eliminating the errors caused by specular reflections.

Rodr\'iguez \& Rubio~\cite{rodriguez2005new} used two strategies to locate inner and outer iris contours. 
For locating the inner contour of the iris, the operator proposed by Daugman is used. 
Then, for determining the outer boundary of the iris, three points are detected, which represent the vertexes of a triangle inscribed in a circumference that models the iris boundary. This approach presented no better accuracy than the Daugman method, but makes full use of the local texture variation and does not use any optimization procedure. 
For this reason, it can reduce the computational cost~\cite{rodriguez2005new}. 

Alvarez-Betancourt \& Garcia-Silvente \cite{alvarez2010fast} presented an iris location method based on the detection of circular boundaries under an approach of gradient analysis in points of interest of successive arcs.
The quantified majority operator QMA-OWA~\cite{pelaez2006majority} was used in order to obtain a representative value for each successive arc. 
The identification of the iris boundary is given by obtaining the arc with the greatest representative value. The authors reported similar results to those achieved by the Daugman method, with improvements in processing time.

In the method proposed by ZhuYu \& Cui~\cite{cui2012rapid}, the first step is to remove the eyelashes by dual-threshold method, which can be an advantage over other iris location approaches. 
Next, the facula is removed through erosion method. 
Finally, the accurate location is obtained through Hough Transform and least-squares method.

Zhou et al.~\cite{zhou2013new} presented a method for iris location based on Vector Field Convolution (VFC), which is used to estimate the initial location of the iris. This initial estimate makes pupil location much closer to the real boundary instead of circle  fitting, improving location accuracy and reducing computational cost.
The final result is obtained using the algorithm proposed by Daugman~\cite{daugman2004iris}.

Zhang et al.~\cite{zhang2014new} used an algorithm which adopts a momentum based level set method~\cite{lathen2009momentum, wang2010momentum} to locate the pupil boundary. 
Finally, the Daugman's method was used in order to locate the iris.
Determine the initial contour for momentum based level set by minimum average gray level method decreases the time consumption and improves the results obtained by the Daugman's method. This improvement happens because this initial contour, as well as the Zhou et al.~\cite{zhou2013new} approach, is generally close to the real iris inner boundary~\cite{zhang2014new}.

Su et al.~\cite{su2017iris} proposed an iris location algorithm based on regional property and iterative searching. 
The pupil area is extracted using the regional attribute of the iris image, and the iris inner edge is fitted by iterating, comparing and sorting the pupil edge points. The outer edge location is completed in an iterative searching method on the basis of the extracted pupil centre and radius.

As can be seen, several works in the literature have proposed methods to perform iris location by determining a circle that delimits it (as shown in \textit{red} in Figure~\ref{fig:iris}), since in many applications it is necessary to perform the iris normalization.
Normalization consists in transforming the circular region of the iris from the Cartesian space into a polar coordinate system, so that the iris is represented by a rectangle.
Usually, representations and characteristics used on further processes are extracted from the transformed image.

In contrast, with the increasing success of deep learning techniques and \glspl*{cnn} in computer vision problems~\cite{Hinton:2006,Pinto:2011a,Krizhevsky12, menotti2015deep, mao2017eeg, fan2017multi}, it has become interesting also in iris-related biometrics problems (besides faces) the use of the entire iris region, including the pupil and some sclera region, without the need for normalization.

In this sense, this work defines the iris location task as the determination of the smallest \textit{squared} bounding box that encompasses the entire region of the iris as show in \textit{yellow} in Figure~\ref{fig:iris}.
Thus we propose to evaluate, as baselines, the following window-based detectors:
1) a sliding window detector based on features from \gls*{hog} and a linear \gls*{svm} classifier, i.e., an adaptation from the human detection method proposed by Dalal \& Triggs~\cite{dalal2005histograms};
2) a deep learning based detector fine-tuned from YOLO object detector~\cite{redmon:2016,redmon:2017}.

We compare our results with the well-known method of Daugman~\cite{daugman1993high}, since its notoriety and one fair implementation can be publicly found\footnote{\url{https://github.com/Qingbao/iris}}.
The experiments were performed in six databases and the reported results show that the use of deep learning to iris location is promising. 
The fine-tuned model from YOLO object detector yielded real-time location with high accuracy, overcoming problems such as noise, eyelids, eyelashes and reflections.

This paper is structured as follows: 
Section~\ref{sec:datasets} presents the databases used in the experiments;
Section~\ref{sec:baselines} describes the baseline methods used in this work;
Section~\ref{sec:results} reports our experiments and discusses our results;
Finally, Section~\ref{sec:conclusion} concludes the work.

\section{Databases}
\label{sec:datasets}

Six databases were used for the experiments performed in this work: \gls*{iiitdcli}~\cite{kohli2013revisiting}, \gls*{ndcld15}~\cite{doyle2015robust}, \acrshort*{mobbiofake}~\cite{sequeira2014iris}, \gls*{ndccl}~\cite{doyle2014notre}, CASIA-IrisV3 Interval~\cite{interval} and BERC mobile-iris database~\cite{kim2016empirical}. 

Except the \gls*{ndcld15}, all other databases were manually annotated from a single annotator\footnote{The iris location annotations are publicly available to the research community at \url{https://web.inf.ufpr.br/vri/databases/iris-location-annotations/}}.
The \gls*{ndcld15} annotations were provided by the database authors~\cite{doyle2015robust}.

Bellow we present a brief description of these databases and how they were used in the experiments.

\textit{\textbf{\acrlong*{iiitdcli}:}} The \gls*{iiitdcli} database consists of $6570$ iris images of $101$ individuals.
Three classes of images were used for the composition of the database: individuals who are not using contact lenses, individuals using transparent lenses and individuals using color cosmetic lenses. In order to study the effect of the acquisition device, iris images were captured using two sensors: Cogent iris sensor and VistaFA2E single iris sensor~\cite{kohli2013revisiting}. 

\begin{figure}[!htb]
	\vspace{-2.5mm} 
	\centering
	\subfloat[][\centering \gls*{iiitdcli} (VistaFA2E~sensor)\label{fig:vista}]{%
		\includegraphics[width=0.31\columnwidth]{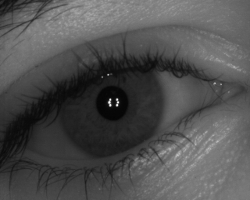}} \,
    \subfloat[][\centering \gls*{iiitdcli} (Cogent~sensor)\label{fig:cogent}]{%
		\includegraphics[width=0.31\columnwidth]{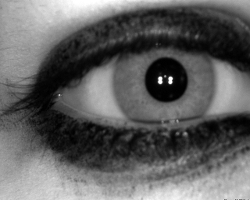}} \,
	\subfloat[][\centering BERC \label{fig:berc}]{%
		\includegraphics[width=0.31\columnwidth]{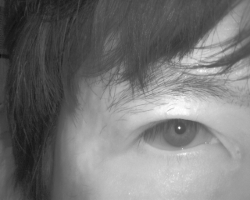}} \\[-1ex]
		
	\subfloat[][\centering MobBIO (Fake)\label{fig:fake}]{%
		\includegraphics[width=0.31\columnwidth]{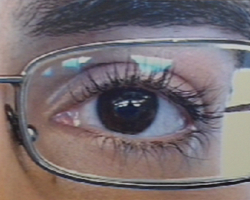}} \,
    \subfloat[][\centering MobBIO (Real)\label{fig:real}]{%
		\includegraphics[width=0.31\columnwidth]{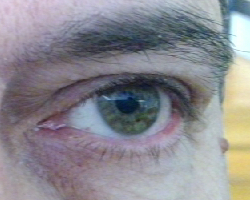}} \,
	\subfloat[][\centering CASIA-IrisV3 Interval\label{fig:interval}]{%
		\includegraphics[width=0.31\columnwidth]{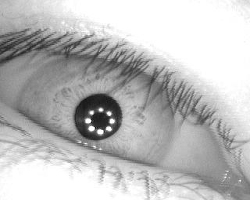}} \\[-1ex]
		
	\subfloat[][\centering \gls*{ndccl} (AD100~sensor)\label{fig:ad100}]{%
		\includegraphics[width=0.31\columnwidth]{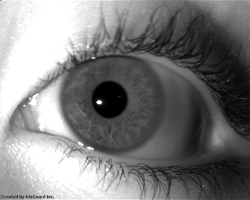}} \,
    \subfloat[][\centering \gls*{ndccl} (LG4000~sensor)\label{fig:lg4000}]{%
		\includegraphics[width=0.31\columnwidth]{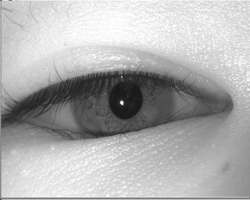}} \,
	\subfloat[][\centering \gls*{ndcld15}\label{fig:ndcld}]{%
		\includegraphics[width=0.31\columnwidth]{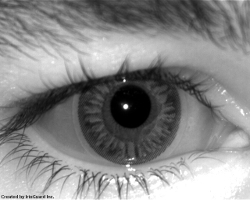}}
		
	\caption{Examples of images from the databases used.} 

\label{fig:examplesdatasets} 
\end{figure}

For the training set, $1500$ images of each sensor were randomly selected. 
The remaining images ($3570$) were used to compose the test set. 
All images have resolution of $640\times480$ pixels and were manually annotated. 
Figure~\ref{fig:vista} and Figure~\ref{fig:cogent} show, respectively, examples of images obtained by VistaFA2E and Cogent sensors.

\textit{\textbf{CASIA-IrisV3 Interval}} - This database consists of $2639$ iris images with resolution of $320\times280$ pixels, obtained in two sections. The images were captured with their own developed camera and an example can be seen in Figure~\ref{fig:interval}.
The main characteristic of this database is that a circular near-infrared led illumination was used when the images were captured, thus this database can be used for studies on the detailing of texture features in iris images~\cite{interval}.
For training, $1500$ images were randomly selected. The remaining images were used for testing.

\textit{\textbf{\acrlong*{ndccl}}} - The images from the \gls*{ndccl} database have resolution of $640\times480$ pixels and were captured under near-infrared illumination. Two iris cameras were used: IrisGuard AD100 (Figure~\ref{fig:ad100} and IrisAccess LG4000 sensor (Figure~\ref{fig:lg4000}), composing two subsets. The IrisAccess LG4000 subset has a training set with $3000$ images and a test set of $1200$ images. IrisGuard AD100 subset has $600$ images for training and $300$ for testing~\cite{doyle2013variation, yadav2014unraveling}.
The database contains images of individuals divided into three classes: no contact lenses, non-textured contact lenses and textured contact lenses.

\textit{\textbf{\acrshort*{mobbiofake}}} - The \acrshort*{mobbiofake} database was created with the purpose of studying the liveliness detection in iris images obtained from mobile devices in uncontrolled environments~\cite{sequeira2014iris}. This database is composed of $1600$ fake iris images of $250\times200$ pixels, obtained from a subset of $800$ images belonging to the MobBIO database \cite{sequeira2014mobbio}. 

For the creation of the fake images, the original images were grouped by each subject and a pre-processing was performed in order to improve the contrast. The images were then printed using a professional printer in a high quality photo paper and recaptured using the same device. Finally, the images were cropped and resized to unify the dimensions. The database is equally divided into training and test sets, in other words, $400$ real images and $400$ fake images were destined for the training sets. Figure~\ref{fig:fake} and Figure~\ref{fig:real} are examples of fake and real images, respectively.

\textit{\textbf{Notre Dame Contact Lens Detection 2015}} - The \gls*{ndcld15} database is composed of $7300$ iris images with resolution of $640\times480$ pixels. 
This database is composed of $6000$ images for training and $1300$ images for evaluation. 
Images were acquired using either IrisAccess LG4000 sensor or Iris-Guard AD100 sensor.
All iris images were captured in a windowless indoor lab under consistent lighting conditions.
This database was created with the purpose of studying the classification of iris images between types of contact lenses~\cite{doyle2015robust}. 
Therefore, the database contains images of individuals divided into three classes: no contact lenses, non-textured contact lenses and textured contact lenses. 
An example image of this database can be seen in Figure~\ref{fig:ndcld}.

\textit{\textbf{BERC Mobile-iris Database}} - The BERC database is composed of images obtained in near-infrared wavelength with a resolution of $1280\times960$ pixels. 
The images were captured by a mobile device under vertical position, in sequences composed of $90$ images~\cite{kim2016empirical}.
In order to simulate the situation where the user moves the mobile phone back and forth to adjust the focus, the sequences of images were obtained by moving the mobile phone to the iris at $3$ distances: $40$ to $15$ cm, $15$ to $25$ cm and $25$ to $15$ cm.
The best images of each sequence were selected, totaling $500$ iris images of $100$ subjects.
An example image of this database can be seen in Figure~\ref{fig:berc}.
In this database, $400$ images were randomly selected for training and $100$ for testing.

\section{Baselines}
\label{sec:baselines}

In this work, we use two approaches to perform iris location. 
One of them is based on \gls*{hog} and \gls*{svm}, which is an adaptation of the human detection method proposed by Dalal \& Triggs~\cite{dalal2005histograms}. 
We use this approach together with the sliding window technique presented on the face detection method, proposed by Viola \& Jones~\cite{viola2001}, \cite{viola2004}.
The other approach is based on deep learning, using YOLO \glspl*{cnn}~\cite{redmon:2016}.

\subsection{Histogram of Oriented Gradients and Support Vector Machines}
Despite image acquisition with different devices, lighting conditions, variations of translation, rotation and scale~\cite{zhu2000biometric}, the iris presents a common structure, following patterns of texture, shape and edge orientations, which can be described by a feature descriptor and interpreted by a classifier.

\begin{figure}[!htb]
\centering
\includegraphics[width=\linewidth]{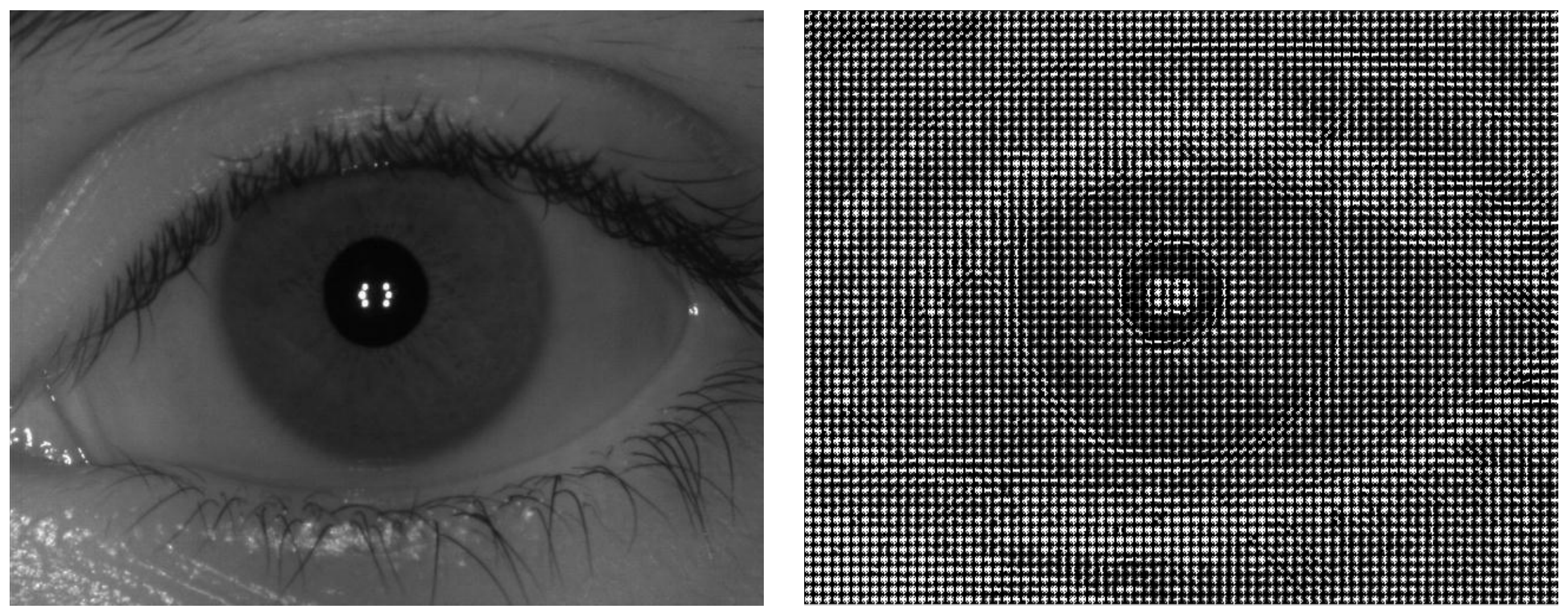}
\vspace{-6mm}
\caption{Exemple of image described by \gls*{hog}.}
\label{fig:hog}
\vspace{-2mm}
\end{figure}

\gls*{hog} is a feature descriptor used in computer vision for object detection. This method quantizes the gradient orientation occurrences in regions of an image, extracting shape information from objects~\cite{dalal2005histograms}. 
Figure~\ref{fig:hog} illustrates an image described by \gls*{hog}.

In this work, each window was divided into cells of $8\times8$ pixels. For each cell, the horizontal and vertical gradients in all pixels are calculated. Thus, the orientations and magnitudes of the gradient are obtained. The gradient orientations are then quantified in nine directions.

In order to avoid effects of light and contrast variation, the histograms of all cells on blocks ($2\times2$ cells) are normalized. The \gls*{hog} feature vector that describes each iris window is then constructed by concatenating the normalized cell histograms for all blocks. Finally, a feature vector ($2\times2$ blocks~$\times$~$8$ cells $\times$~$9$ orientations) is obtained to describe each iris candidate window.

\begin{figure}[!htb]
\subfloat[][\centering Positive samples\label{fig:positives}]{\includegraphics[width=\linewidth]{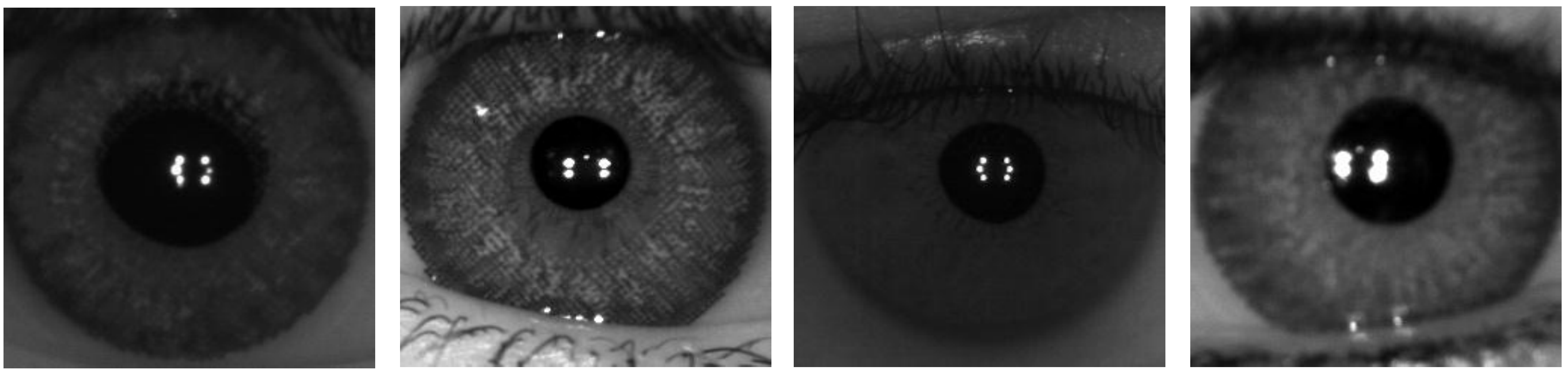}% 
} \\[-1ex]

\subfloat[][\centering Negative samples\label{fig:negatives}]{\includegraphics[width=\linewidth]{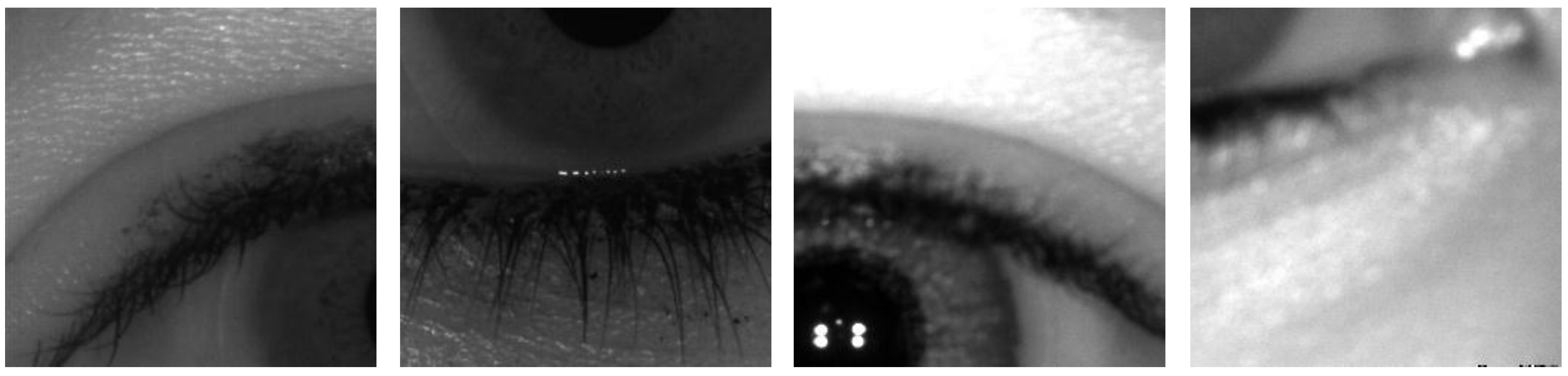}%
}
\caption{Training samples used by \gls*{svm}.}
\end{figure}

The window containing the iris region (ground truth) from each training image is extracted and used to compose the examples of positive windows.
Furthermore, windows that are completely outside or have only a small intersection with the iris region are extracted and considered negative windows. We created $10$ negative windows for each positive window.
Figures~\ref {fig:positives} and \ref{fig:negatives} illustrate, respectively, positive and negative samples used for the training of the proposed approach.

From these positive and negative samples, the SVM classifier is trained using a linear kernel and the constant is determined by grid-search in the training set.

The \gls*{svm} was first presented by Vladimir Vapknik~\cite{boser1992training}, and is one of the most used classification methods in recent years~\cite{franchi2016hyperspectral,ruiz2014bayesian}. 
To find the decision boundary, the \gls*{svm} minimizes the upper limit of the generalization error, which is obtained by maximizing the margin distance from the training data.

In order to perform the iris location, a sliding window approach with different scales is applied in each test image.
We adopted windows with size $50\times50$ pixels as canonical scale. From this scale, we used $6$ lower scales and $8$ higher scales by a factor of $5\%$.
The image region that presents the greatest similarity with the iris can be found through the decision border generated by the SVM, which will return the highest positive response for the best estimated iris location.

\subsection{YOLO Object Detector}
Currently, deep \glspl*{cnn} are one of the most efficient ways to perform image classification, segmentation and object detection.  
In this work, we use the Darknet~\cite{darknet13}, which is an open source neural network framework used to implement YOLO, a state-of-the-art real-time object detection system~\cite{redmon:2016}. 

The YOLO network, as most CNNs, is composed of three main operation layers to object detection, which are: convolution, max pooling and classification, the latter occurs through fully connected layers.

On Darknet, convolutional layers work as feature extraction, in other words, a convolutional kernel is sliding in the input image.
The network architecture is inspired by the GoogLeNet model for image classification~\cite{szegedy2015going}. 
The original YOLO has $24$ convolutional layers that produce different feature maps from the input. 

The feature maps are then processed by max pooling layers, which dimensionally reduces the previously obtained feature map.  max pooling divides the feature map into blocks and reduces each block into one value.
Instead of the inception modules used by GoogLeNet, YOLO uses $1 \times 1$ reduction layers followed by $3 \times 3$ convolutional layers, similar to Lin et al.~\cite{lin2013network}.

However, in this work we use an fast version of YOLO, based on a neural network with fewer convolutional layers ($9$ instead of $24$) and fewer filters in those layers. 
Other than the size of the network, all training and testing parameters are the same for both YOLO and Fast-YOLO.

\section{Results and discussion}
\label{sec:results}

In this work, we evaluate both \gls*{hog}-\gls*{svm} and YOLO approaches, applied to iris location and compare them to the well-known Daugman method. 
The experiments were performed in the six databases described in Section~\ref{sec:datasets}. 
All experiments were performed on a NVIDIA Titan XP GPU ($3840$ CUDA cores and $12$ GB of RAM) and also using an Intel (R) Core i$7$-$5820$K CPU @ $3.30$GHz $12$ core, $64$GB of DDR$4$ RAM.

In order to analyze the experiments, we employ the following metrics: Recall, Precision, Accuracy and \gls*{iou}. 
These metrics are defined between the area of the ground truth and predicted bounding boxes in terms of False Positives (FP), False Negatives (FN), True Positives (TP), and True Negatives (TN) pixels, and can formally be expressed as:

\begin{eqnarray*}
\textit{Recall }              &=& \frac{TP}{TP+FN},\\ \\ 
\textit{Precision}            &=& \frac{TP}{TP+FP},\\ \\ 
\textit{Accuracy}             &=& \frac{TP+TN}{TP+TN+FP+FN},\\ \\ 
\textit{IoU}   				  &=& \frac{TP}{FP+TP+FN} \\
\end{eqnarray*}

In the following, we describe experiments in three different scenarios: intra-sensor, inter-sensor, multiple-sensors and mixing of databases.

\begin{table*}[!ht]
	\centering
	\caption{Intra-sensor results (\%)}
	\label{tab:intra}
	\begin{tabular}{c c c c c c c c c c c c c}\hline  \\[\dimexpr-\normalbaselineskip+4pt]

		 \multirow{3}{*}{\textbf{Database}}  & \multicolumn{3}{c}{\textbf{Recall}} & \multicolumn{3}{c}{\textbf{Precision}} & \multicolumn{3}{c}{\textbf{Accuracy}} & \multicolumn{3}{c}{\textbf{IoU}} \\
		 \\[\dimexpr-\normalbaselineskip+2pt] 
		 
& Daugman & HOG & \multirow{2}{*}{YOLO}  
 & Daugman & HOG & \multirow{2}{*}{YOLO} 
  & Daugman & HOG & \multirow{2}{*}{YOLO} 
   & Daugman & HOG & \multirow{2}{*}{YOLO} \\
& ~\cite{daugman2004iris} 
             & SVM &    
 & ~\cite{daugman2004iris} 
              & SVM &    
  & ~\cite{daugman2004iris} 
               & SVM &    
   & ~\cite{daugman2004iris} 
                & SVM &     \\
		 \\[\dimexpr-\normalbaselineskip+2pt] \toprule
		 \\[\dimexpr-\normalbaselineskip+2pt]	
		\textbf{\gls*{ndccl}} \\ AD100     &  84.60  &  92.39  & \textbf{98.78} & 82.49   & 94.78    & \textbf{95.03} & 94.28   & 96.98   & \textbf{98.49}  & 80.41   & 87.52      & \textbf{93.84}   \\ 
		        LG4000             &  93.41  & 96.72   & \textbf{97.81} & 92.15   & 90.80    & \textbf{97.73} & 97.53   & 97.24   & \textbf{99.05}  & 89.67   & 87.76      & \textbf{95.06}   \\ \midrule
		\textbf{\gls*{iiitdcli}} \\  Vista &  85.49  & 94.51   & \textbf{97.85} & 89.34   & 92.24    & \textbf{93.71} & 95.38   & 98.10   & \textbf{98.28}  & 80.82   & 87.23      & \textbf{91.76}   \\ 
		 Cogent                      &  86.24  & \textbf{96.44}   & 96.02 & 92.82   & 87.99    & \textbf{95.58} & 96.34   & 96.67   & \textbf{98.33}  & 82.61   & 84.76      & \textbf{91.84}   \\ \midrule
		
		\textbf{MobBIO} \\ Real    &  76.32  & 95.77   & \textbf{96.81} & 74.71   & 72.26    & \textbf{94.02} & 85.26   & 95.33   & \textbf{98.97}  & 70.79   & 68.76      & \textbf{91.02}   \\ 
		                   Fake    &  75.81  & 93.28   & \textbf{96.06} & 73.45   & 74.33    & \textbf{95.05} & 84.81   & 95.26   & \textbf{98.90}  & 70.12   & 68.99      & \textbf{91.27}   \\ \midrule
		                   \textbf{BERC}              &  88.19  & 92.83   & \textbf{98.10} & 85.64   & 87.95    & \textbf{93.56} & 98.72   & 98.49   & \textbf{99.71}  & 79.10   & 85.10      & \textbf{91.15}   \\ \midrule
		\textbf{CASIA IrisV3} \\
		     Interval              &  96.38  & 96.97   & \textbf{97.79} & \textbf{96.23}   & 88.48    & 96.02 & \textbf{97.38}   & 92.21   & 97.10  & 90.95   & 86.17      & \textbf{91.24}   \\ \midrule
		\textbf{\gls*{ndcld15}}           &   91.63 & 96.04   & \textbf{97.28} & 89.76   & 90.29    & \textbf{95.71} & 96.67   & 97.14   & \textbf{98.54}  & 85.34   & 86.85      & \textbf{93.25}   \\ \bottomrule
		
	\end{tabular}
\end{table*}

\textbf{Intra-sensor}: 
Table~\ref{tab:intra} shows the results obtained by intra-sensor experiments, in other words, experiments in which the models were trained and tested with images from the same sensor.
The YOLO \gls*{cnn} achieved the best averages in almost all analyzed metrics and required less processing time for iris location per image. 
The exception is for CASIA IrisV3 Interval database where Daugman method presented slightly better Precision ($96.23\%$ against $96.02\%$) and Accuracy ($97.38\%$ against $97.10\%$).
This surprising result can be explained by the high level of cooperation and control in the image acquisition of such database. 
That is, the Daugman method take somehow advantage of the scenario.
Anyway, the YOLO \gls*{cnn} locates the iris in real-time ($0.02$ seconds per image, on average) using our fast Titan XP GPU, whilst the Daugman method and the HOG-SVM approach demand, on average, $3.5$ and $5.2$ seconds, respectively, to locate the iris in each image using a single CPU core.

\begin{table*}[!ht]
	\centering
	\caption{Inter-sensor results (\%)}
	\label{tab:inter}
	\begin{tabular}{c c c c c c c c c c c}\hline  \\[\dimexpr-\normalbaselineskip+4pt]

		 \multirow{2}{*}{\textbf{Database}} & \multicolumn{2}{c}{\textbf{Set}} & \multicolumn{2}{c}{\textbf{Recall}} & \multicolumn{2}{c}{\textbf{Precision}} & \multicolumn{2}{c}{\textbf{Accuracy}} & \multicolumn{2}{c}{\textbf{IoU}} \\
		 \\[\dimexpr-\normalbaselineskip+2pt] 
		 
		          & Train & Test & HOG-SVM & YOLO & HOG-SVM & YOLO & HOG-SVM & YOLO & HOG-SVM & YOLO \\ \toprule

		\multirow{2}{*}{\textbf{\gls*{ndccl}}}     & AD100 & LG4000  & \textbf{92.95}  & 79.25 & \textbf{91.13} & 89.18 & \textbf{96.84} & 92.67 & \textbf{85.78} & 68.71 \\ 
		                   & LG4000 & AD100  & 93.22  & \textbf{97.99} & 93.15 & \textbf{93.59} & 96.78 & \textbf{97.94} & 86.76 & \textbf{91.63}  \\ \midrule
		\multirow{2}{*}{\textbf{\gls*{iiitdcli}}} & Vista & Cogent  & \textbf{96.89}  & 96.13 & 89.89 & \textbf{94.21} & 96.43 & \textbf{97.98} & 83.94 & \textbf{90.57} \\ 
		                   & Cogent & Vista  & 93.44  & \textbf{98.26} & \textbf{93.61} & 87.97 & \textbf{97.08} & 96.65 & \textbf{87.55} & 80.92  \\ \bottomrule
	\end{tabular}
\end{table*}

\begin{figure}[!htb]
\centering
\subfloat[]{\includegraphics[width=\linewidth]{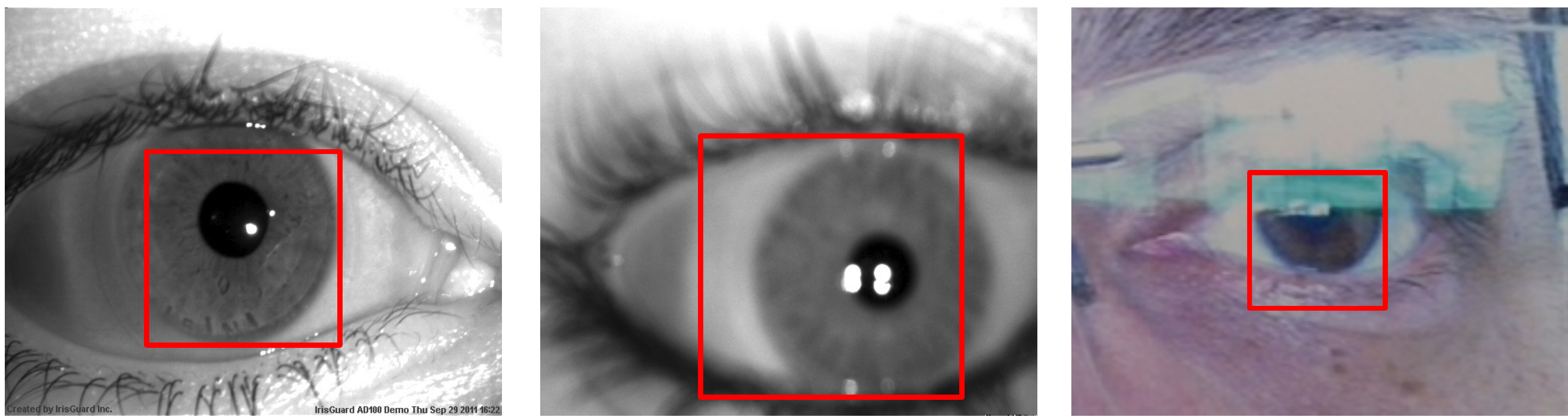}%
\label{fig:poor_results}} \vspace{-2mm}

\subfloat[]{\includegraphics[width=\linewidth]{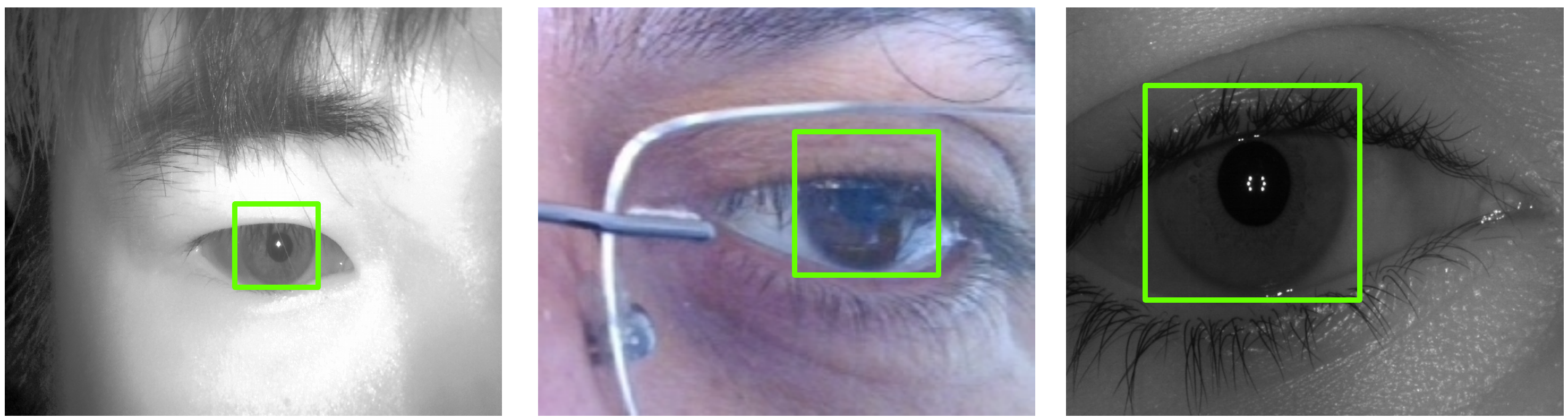}%
\label{fig:good_results}}
\caption{Samples of iris location obtained in the experiments: (a) poor results due to a homogeneous training set; (b) good results achieved with images of different sensors on training set.}
\end{figure}

\textbf{Inter-sensor}: 
In addition, for databases containing images acquired with more than one sensor, inter-sensor experiments were performed and are presented in Table~\ref{tab:inter}. 
That is, we train the detectors with images of one sensor and test/evaluate then on the images from other sensor.
These experiments show that in some cases YOLO \gls*{cnn} did not achieve promising results as previously shown.
For example, in the database \gls*{ndccl}, when fine tunning/training the detector with images from the AD100 sensor and testing with the ones from LG4000 sensor.
The reason for the poor result might lie in the fact that the database for that specific sensor (AD100) has only $600$ images, thus not allowing a good generalization of the trained CNN. 
In Figure~\ref{fig:poor_results}, we can observe some examples where the iris location obtained by the YOLO method did not achieved good results. 

\begin{table*}[!ht]
	\centering
	\caption{Combined sensor results (\%), same databases}
	\label{tab:combined}
	\begin{tabular}{c c c c c c c c c c c}\hline  \\[\dimexpr-\normalbaselineskip+4pt]

		 \multirow{2}{*}{\textbf{Database}} & \multicolumn{2}{c}{\textbf{Set}} & \multicolumn{2}{c}{\textbf{Recall}} & \multicolumn{2}{c}{\textbf{Precision}} & \multicolumn{2}{c}{\textbf{Accuracy}} & \multicolumn{2}{c}{\textbf{IoU}} \\
		 \\[\dimexpr-\normalbaselineskip+2pt] 
		 
		          & Train & Test  & HOG-SVM & YOLO & HOG-SVM & YOLO & HOG-SVM & YOLO & HOG-SVM & YOLO \\ \toprule

		\multirow{2}{*}{\textbf{\gls*{ndccl}}}     & AD100 \& LG4000 & LG4000  & 95.37  & \textbf{99.29} & 92.93 & \textbf{99.68} & 97.48 & \textbf{99.77} & 88.63 & \textbf{98.91}  \\ 
		                   & AD100 \& LG4000 & AD100   & 91.77  & \textbf{99.37} & 94.77 & \textbf{97.42} & 96.85 & \textbf{99.36} & 86.91 & \textbf{96.85}  \\ \midrule
		\multirow{2}{*}{\textbf{\gls*{iiitdcli}}} & Vista \& Cogent & Cogent  & 96.73  & \textbf{97.26} & 87.15 & \textbf{96.48} & 96.50 & \textbf{98.49} & 84.17 & \textbf{92.50}  \\ 
		                   & Vista \& Cogent & Vista   & 94.20  & \textbf{98.34} & 92.74 & \textbf{93.79} & 97.01 & \textbf{98.55} & 87.41 & \textbf{91.78}  \\ \bottomrule
		
	\end{tabular}
\end{table*}

\textbf{Multiple-sensors}: 
In order to better analyze and understand the results of the inter-sensor experiments and to confirm our hypothesis that the YOLO's poor performance is due to few/homogeneous training samples, experiments were performed combining images from multiple sensors of the same databases. 
The figures obtained in this new experiment can be seen in Table~\ref{tab:combined}.
It highlights the importance of a diverse collection of images for the training set in \glspl*{cnn}. 
With a larger number of images acquired from different sensors in the training set, the CNN was able to better generalize, increasing the correct iris location in most cases. 
Some examples of good iris location can be seen in Figure~\ref{fig:good_results}.

\begin{table*}[!ht]
	\centering
	\caption{Combined sensor results (\%), mixed databases}
	\label{tab:all}
	\begin{tabular}{ c c c c c c c c}\hline  \\[\dimexpr-\normalbaselineskip+4pt]

		 \multirow{2}{*}{\textbf{Method}} & \multicolumn{2}{c}{\textbf{Set}} & \multirow{2}{*}{\textbf{Recall}} & \multirow{2}{*}{\textbf{Precision}} & \multirow{2}{*}{\textbf{Accuracy}} & \multirow{2}{*}{\textbf{IoU}} & \multirow{2}{*}{\textbf{Time}} \\
		 \\[\dimexpr-\normalbaselineskip+2pt] 
		 
		          & Train & Test    &  &  &  & &  \\ \toprule

		YOLO     & All training sets & All test sets    & \textbf{97.13}  & \textbf{95.20} & \textbf{98.32} & \textbf{92.54} &  \textbf{0.02 s}  \\ 
		                   
		Daugman~\cite{daugman2004iris} & - & All test sets   & 86.45  & 86.28 & 94.04 & 81.09 & 3.50 s   \\ \bottomrule
		                   		
	\end{tabular}
\end{table*}

\begin{figure}[!htb]
\centering
\includegraphics[width=\linewidth]{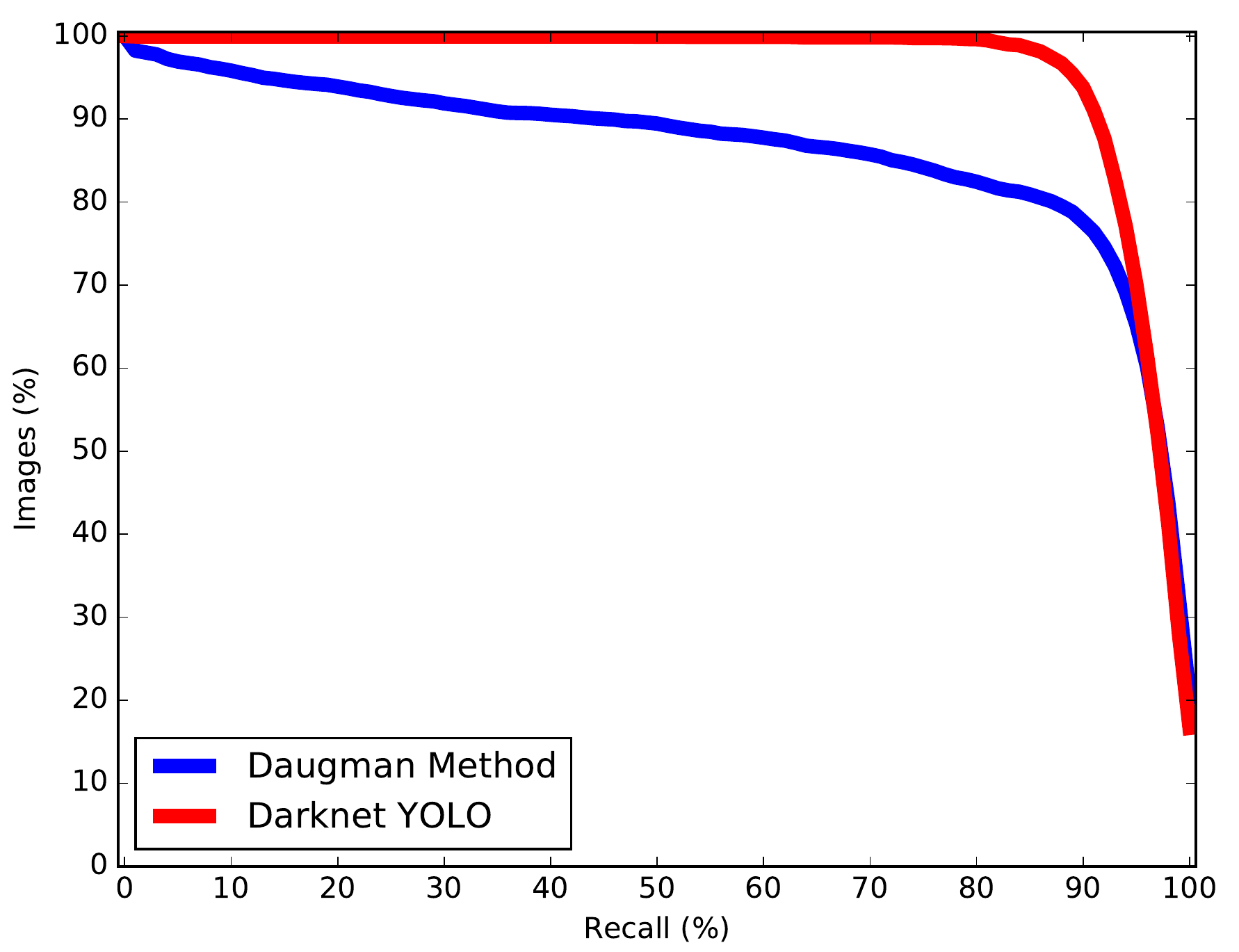}
\caption{Recall curve of both Daugman and YOLO methods applied to all test sets.}
\label{fig:graphic_recall}
\end{figure}

\textbf{Mixing databases}:
Table~\ref{tab:all} contains the results obtained by experiments where YOLO was trained with the training sets of all the databases and tested in the test images of all the databases.
The results achieved by the Daugman method applied to all the test images are also presented, and we used specific parameters for each database.
By analyzing these figures, we observe that YOLO strikingly outperforms the Daugman method in all analyzed metrics.

Figure~\ref{fig:graphic_recall} shows the behavior of the recall curve for the experiment reported in Table~\ref{tab:all}. 
It depicts how the percentage of images varies when we required a minimum Recall rate.
These curve highlights how YOLO is a promising alternative to iris location, since all tested images achieved Recall values above $80\%$.
That is, at least $80\%$ of the required region of a iris is certainly located by the YOLO detector.

\section{Conclusion}
\label{sec:conclusion}

The iris location is a preliminary but extremely important task in specific applications such as iris recognition, spoofing and liveness detection, as well as contact lens detection, among others.
In this work, two object detection approaches were evaluated for the iris location.
The experiments were performed in six databases. 
We manually annotated four of the six databases used in this work, and those annotations are publicly available to the research community.

The experiments showed that the use of the YOLO object detector, based on deep learning, applied to the iris location presents promising results for all studied databases. 
Moreover, the iris location using this approach runs in real-time ($0.02$ seconds per image, on average) using a current and powerful GPU (NVIDIA GeForce Titan XP Pascal).
Another relevant conclusion to be mentioned is that, similar to other deep learning approaches, it is important to have a sufficiently large number of images for training.
The number and variety of images in the training set directly affects the generalization capability of the learned model.

As future work, we intend to perform experiments with more visible and cross-spectral iris databases. 
In addition, we intend to analyze the impact that iris location exerts on iris recognition, spoofing, liveness, and contact lens detection systems.
Also, we plan to study how a short and shallow network than YOLO one can be designed for our single object detection problem, the iris location.

\section*{Acknowledgments}

This research has been supported by Coordination for the Improvement of Higher Education Personnel (CAPES), the Foundation for Research Support of the State of Minas Gerais (Fapemig) and the National Council for Scientific and Technological Development (CNPq) grants 471050/2013-0, \# 428333/2016-8, and \# 313423/2017-2)
We thank the NVIDIA Corporation for the donation of the GeForce GTX Titan XP Pascal GPU used for this research. 
The annotations made in the \gls*{iiitdcli} database are thanks to Pedro Silva (UFOP).

\balance
\bibliographystyle{IEEEtran}
\bibliography{references}

\end{document}